\documentclass[sigconf]{acmart}
\AtBeginDocument{%
  \providecommand\BibTeX{{%
    Bib\TeX}}}

\acmYear{2025} 
\acmConference[HCAI Workshop @ CIKM '25]{Human-Centric AI Workshop}{November 14, 2025}{Seoul, Republic of Korea}

\newcommand{\BibTeX}{B\kern-.05em{\sc i\kern-.025em b}\kern-.08em\TeX}
\NewDocumentCommand{\model}{o}{%
  \IfNoValueTF{#1}
    {f(\cdot)} 
    {f(#1)}    
}
\newcommand{\params}{\boldsymbol{\theta}}
\renewcommand\footnotetextcopyrightpermission[1]{%
  \footnotetext{%
    {\small © 2025 Copyright held by the owner/author(s).}
  }
}

\usepackage{latexsym}
\usepackage{xparse}

\usepackage{siunitx}    
\usepackage{caption} 
\usepackage{amssymb}
\usepackage{amsmath}
\usepackage{amsthm}
\usepackage{booktabs}
\usepackage{enumitem}
\usepackage{graphicx}
\usepackage{color}
\usepackage{multirow}
\usepackage{algorithm}
\usepackage{algpseudocode}

\newtheorem{definition}{Definition}
\DeclareMathOperator*{\argmin}{arg\,min}

\begin{document}

\title{Enhancing XAI Narratives through Multi-Narrative Refinement and  Knowledge Distillation}
\author{Flavio Giorgi}
\email{giorgi@di.uniroma1.it}
\affiliation{%
  \institution{Sapienza University of Rome}
  \city{Rome}
  \country{Italy}
}

\author{Matteo Silvestri}
\email{silvestri.m@di.uniroma1.it}
\affiliation{%
  \institution{Sapienza University of Rome}
  \city{Rome}
  \country{Italy}
}
\author{Cesare Campagnano}
\email{cesare@pinecone.io}
\affiliation{%
  \institution{Pinecone}
  \country{US}
}
\author{Fabrizio Silvestri}
\email{fsilvestri@diag.uniroma1.it}
\affiliation{%
  \institution{Sapienza University of Rome}
  \city{Rome}
  \country{Italy}
}
\author{Gabriele Tolomei}
\email{tolomei@di.uniroma1.it}
\affiliation{%
  \institution{Sapienza University of Rome}
  \city{Rome}
  \country{Italy}
}

\renewcommand{\shortauthors}{Giorgi et al.}

\begin{abstract}
Explainable Artificial Intelligence has become a crucial area of research, aiming to demystify the decision-making processes of deep learning models. Among various explainability techniques, counterfactual explanations have been proven particularly promising, as they offer insights into model behavior by highlighting minimal changes that would alter a prediction. Despite their potential, these explanations are often complex and technical, making them difficult for non-experts to interpret. To address this challenge, we propose a novel pipeline that leverages Language Models, large and small, to compose narratives for counterfactual explanations. We employ knowledge distillation techniques along with a refining mechanism to enable Small Language Models to perform comparably to their larger counterparts while maintaining robust reasoning abilities. In addition, we introduce a simple but effective evaluation method to assess natural language narratives, designed to verify whether the models' responses are in line with the factual, counterfactual ground truth.  As a result, our proposed pipeline enhances both the reasoning capabilities and practical performance of student models, making them more suitable for real-world use cases. 
\end{abstract}



\maketitle

\makeatletter
\setlength{\footskip}{30pt}
\fancypagestyle{plain}{%
  \fancyfoot{} 
  \fancyfoot[C]{\thepage} 
  \renewcommand{\footrulewidth}{0pt} 
}
\pagestyle{plain}
\thispagestyle{plain}
\makeatother

\section{Introduction}\label{sec:intro}
\captionsetup{skip=2pt}
\begin{figure*}[ht]
    \centering
    \includegraphics[width=0.85\linewidth]{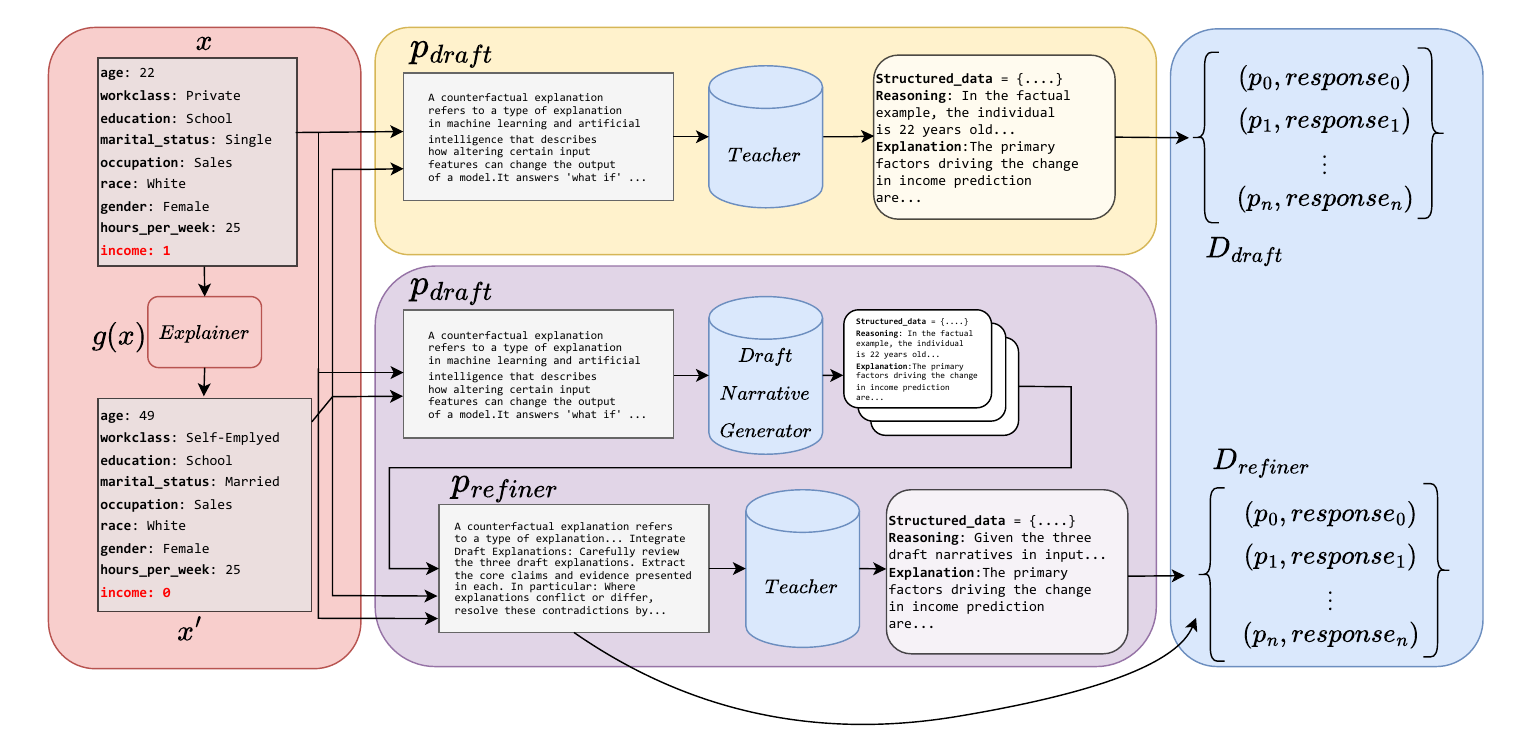}
    \caption{\small Overview of the dataset generation process for the draft narrative generation step and the refiner step using knowledge distillation. Given a factual instance $\boldsymbol{x}$ with its features (e.g., age, work class, education, etc.), a counterfactual generator $g(\boldsymbol{x})$ modifies the instance to create a counterfactual example $\boldsymbol{x}'$, altering specific attributes (e.g., age, work class) to yield a different predicted outcome (e.g., income = 0) (red zone). The factual and counterfactual instances are then integrated into the prompt that is fed into the teacher model, which produces the narrative for the draft narrative generation step (yellow zone) and for the refiner step (purple zone). For the draft narrative generation step, the response is collected and put along with the prompt $p_{draft}$ in the new dataset $D_{draft}$. Concerning the dataset for the refiner step, we generate $N=3$ draft responses and integrate all three into the prompt $p_{refiner}$. Finally, we feed the prompt into the teacher, collect the response, and put the newly formed pair $p_{refiner}, response$ into the dataset $D_{refiner}$.}

    \label{fig:dataset-creation}
\end{figure*}
Counterfactual Explanations (CE) have become a cornerstone of Explainable AI (XAI), providing intuitive insights into model decisions by demonstrating how minimal changes to input features can alter outcomes. This approach is increasingly critical in domains like healthcare, finance, and policy-making, where transparency is essential for trust, accountability, and compliance with emerging AI policies. For instance, the European Union's AI Act~\citep{eu2024ai} mandates explainability for high-risk AI systems, while U.S. guidelines on AI governance~\citep{ostp2023ai} emphasize interpretable decision-making to mitigate bias and ensure fairness. While early XAI efforts targeted specific data types—such as graph-based models in drug discovery~\citep{he2024explaining}, recent work has expanded to diverse modalities, including tabular data and natural language. Large Language Models (LLMs) have been instrumental in this shift, converting technical counterfactuals into human-readable narratives. Fredes and Vitri\`{a}~\citep{fredes2024using} showcase LLM-driven explanations for tabular data counterfactuals, prioritizing user accessibility, while Cheng et al.~\citep{cheng2024llms} explore LLMs' role in generating and evaluating counterfactuals for natural language tasks. Similarly, He et al.~\citep{he2024explaining} and Wang et al.~\citep{wang2024graph} demonstrate LLMs' versatility in graph-based contexts, suggesting their potential as a unifying tool across XAI applications.

Building counterfactual narratives, particularly for tabular data, which underpins applications like credit scoring and medical diagnostics, is extremely important to improve models' explainability. Current methods often remain domain-specific or computationally expensive, limiting their scalability and alignment with policy-driven transparency requirements~\citep{eu2024ai, ostp2023ai}. 
This paper introduces a novel pipeline for counterfactual narrative generation, centering on tabular data. We propose a methodology leveraging Language Models to produce semantically rich, user-centric counterfactual narratives.
By aligning with AI policy mandates for interpretability, this work advances the practical adoption of counterfactual reasoning in XAI, enhancing model transparency and usability in real-world, regulated environments. Our contributions can be summarized as follows:
\begin{itemize}
    \item We introduce a novel pipeline for generating coherent counterfactual narratives. 
   \item  We propose a new evaluation framework for assessing the quality of narratives generated by language models.
   \item We released both the code \url{https://github.com/flaat/llm_kd} and the four datasets used to fine-tune the models
\url{https://huggingface.co/datasets/Anon30241/model_kd_llm}.
\end{itemize}
\section{Related Works}
Recent advancements in narrative-driven explainable AI leveraging LLMs significantly enhance the interpretability of complex machine learning models across diverse data types.
This trend addresses a critical gap between the outputs of XAI algorithms and the comprehension of end-users, in particular, non-experts. A significant portion of this research targets tabular data. For instance, Zeng et al.~\citep{zeng2024enhancing} use a locally deployed LLM to translate the numerical outputs of SHAP~\citep{lundberg2017unified} into plain-language summaries, enhancing their accessibility. Similarly, Fredes and Vitrià~\citep{fredes2024using} demonstrate that LLMs can effectively explain sets of counterfactual examples by mimicking human reasoning, achieving high validity on benchmark datasets. Going beyond static explanations, Slack et al.~\citep{slack2022talktomodel} develop TalkToModel, an interactive dialogue system that allows users to converse with a model to understand its predictions.

Researchers also develop broader frameworks for generating and evaluating narrative explanations. Cedro and Martens~\citep{cedro2023tell} propose a comprehensive survey of narrative-driven XAI, introducing "XAIstories" that translate SHAP values into compelling narratives. Recognizing the need for systematic quality control, Zytek et al.~\citep{zytek2024explingo} introduce Explingo, a dual-LLM system composed of a narrator to generate narratives from SHAP explanations and an automated grader to assess them on metrics like accuracy, completeness, and fluency. On the evaluation front, Ichmoukhamedov et al.~\citep{ichmoukhamedov2024good} propose quantitative metrics such as faithfulness and human similarity to standardize the assessment of LLM-generated narratives.

The application of narrative XAI extends beyond tabular data to more complex structures like graphs, images, and 3D point clouds. For Graph Neural Networks (GNNs), several approaches have emerged. Giorgi et al.~\citep{giorgi2025natural}, for example, generate counterfactual explanations for graphs and use open-source LLMs to produce human-readable outputs, while Cedro and Martens’s~\citep{cedro2024graphxain} GraphXAIN translates explanatory subgraphs into natural language, showing high user satisfaction. Pan et al.~\citep{pan2024tagexplainer} develop TAGExplainer to generate faithful and concise explanations for text-attributed graphs, which are common in social networks and recommendation systems. In the visual domain, Castellano et al.~\citep{castellano2024using} combine Grad-CAM heatmaps from image classifiers with LLMs to create textual descriptions that explain why a model focused on certain areas of an artwork. Kočić et al.~\citep{kovcic2025llm} instead present a framework where an LLM is an active participant in the counterfactual generation process itself, selecting segment perturbations in 3D point clouds to ensure they are semantically meaningful.
Finally, these techniques are also being adapted for highly specialized fields. He et al.~\citep{he2024explaining}, for example, apply LLMs to explain GNNs used for molecular property prediction, demonstrating the broad applicability of this paradigm.

Together, these works illustrate a clear and growing trend toward using LLMs to make Machine Learning models more transparent and trustworthy.

\section{Background}
Before introducing our pipeline, we formalize two important definitions, namely, the Counterfactual Explanation Problem (CEP) and the Counterfactual Narrative Generation Problem (CNGP). We take the definition for the CEP from \citep{giorgi2025natural}.

\begin{definition}[The Counterfactual Explanation Problem]\label{def:1}
Given a sample $\boldsymbol{x}$ and a predictive model $\model$ parametrized by $\params$, hereinafter referred to as \textit{oracle}, a distance function $d(\cdot,\cdot)$, and a generic counterfactual generator $g(\cdot)$ the goal is to find a sample $\boldsymbol{x}'$ using $g(\cdot)$ where $\boldsymbol{x}'\neq \boldsymbol{x}$ and $\model[\boldsymbol{x}'] \neq \model[\boldsymbol{x}]$ such that the distance $d(\boldsymbol{x}', \boldsymbol{x})$ is minimized, or $\perp$ if no valid counterfactual example exists. The sample $\boldsymbol{x}'$ is called a \textit{counterfactual example} for $\boldsymbol{x}$. 

\end{definition}

The distance function $d(\cdot, \cdot)$ ensures that the counterfactual sample $\boldsymbol{x}'$ remains as close as possible to the original factual sample $\boldsymbol{x}$.
We can formalize the role of the counterfactual generator $g$ by casting it as the problem of solving the following objective:
\begin{equation*}
\label{eq:cf-train}
\begin{aligned}
\boldsymbol{x}' = \argmin_{\boldsymbol{x^*}}  ~d(\boldsymbol{x}, \boldsymbol{{x^*}}) ~~ \text{ s.t.: } \boldsymbol{x} \neq \boldsymbol{x}^*  \wedge  f(\boldsymbol{x}) \neq f(\boldsymbol{{x^*}}).
\end{aligned}
\end{equation*}

Hereinafter, we assume to have an oracle $\model$ for a binary classification task; the arguments generalize with minor modifications to multiclass classification and regression as well. Below, we formally define the Counterfactual Narrative Generation Problem.

\begin{definition}[The Counterfactual Narrative Generation Problem]\label{def:2}
Given a generic pair of factual-counterfactual samples $(\boldsymbol{x}, \boldsymbol{x}')$ where the counterfactual sample $\boldsymbol{x}'$ is a solution of the Counterfactual Explanation Problem, a generic function $H$, and a prompt $p$ we want to generate a natural language narrative $\varepsilon$ associated with the generic factual-counterfactual pair $(\boldsymbol{x}, \boldsymbol{x}')$, namely, $\varepsilon = H(p, \boldsymbol{x}, \boldsymbol{x}')$ also known as counterfactual narrative.   
\end{definition}
The function $H$ is defined generically to accommodate different instantiations, including deterministic mappings, probabilistic generative models, or ensembles thereof.
\begin{figure*}
    \centering
    \includegraphics[width=1\linewidth]{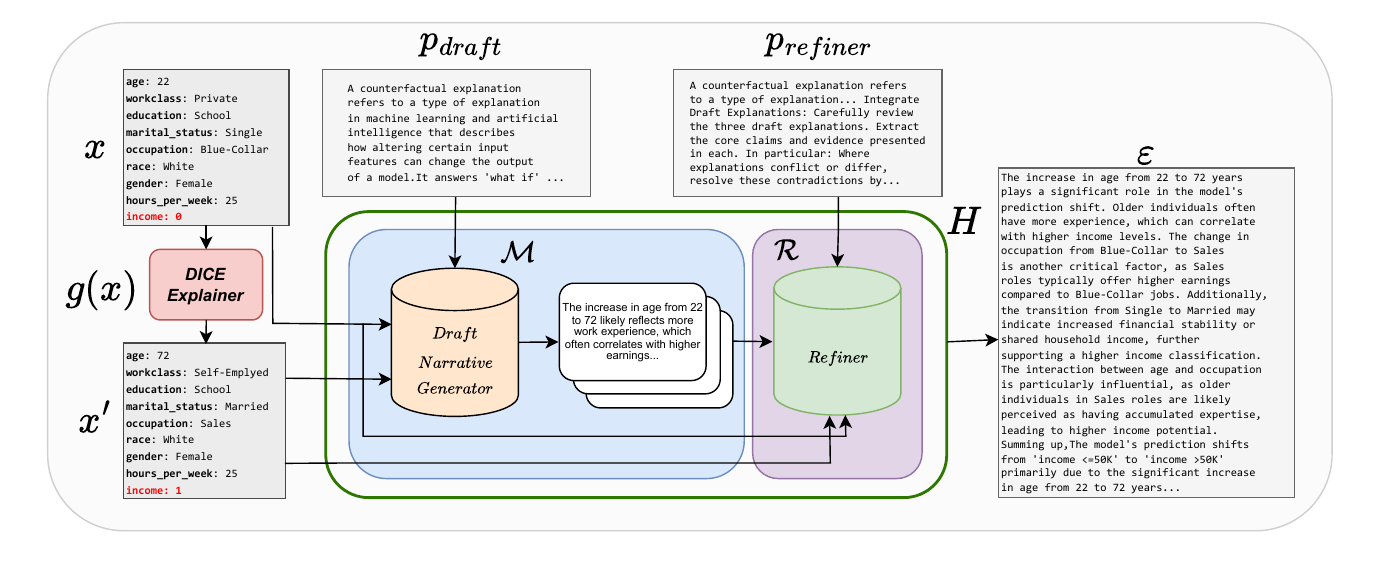}
    \caption{Multi-Narrative Refinement pipeline schema}
    \label{fig:pipeline}
\end{figure*}
\section{Proposed Pipeline}

In order to solve the problem in Definition~\ref{def:2}, we propose a new pipeline (see Figure~\ref{fig:pipeline}, Algorithm~\ref{alg:cap}) to generate coherent and useful narratives for factual-counterfactual pairs. Our approach enhances the capabilities of Small Language Models (SLMs) through a two-stage pipeline called Multi-Narrative Refinement (MNR), which encourages structured self-correction. The entire framework is fine-tuned using knowledge distillation, allowing us to leverage the expertise of a much larger teacher models.

In the first stage, a \emph{Draft Narrative Generator} (DNG) $\mathcal{M}$ is tasked with generating multiple candidate draft explanations. Given a factual instance $\boldsymbol{x}$, its counterfactual $\boldsymbol{x}'$, and a structured prompt $p$ encoding task-specific instructions, the DNG is queried independently $N = 3$ times to produce diverse draft explanations for the prediction shift from $\boldsymbol{x}$ to $\boldsymbol{x}'$. These drafts reflect different plausible reasoning paths the model might follow.

In the second stage, a different model called the \emph{Refiner} $\mathcal{R}$ 
receives as input the same factual instance $\boldsymbol{x}$, counterfactual $\boldsymbol{x}'$, prompt $p$, and the three draft explanations. Its goal is to synthesize a coherent and accurate explanation by comparing, contrasting, and integrating the drafts. This additional refinement step promotes consistency and correctness, encouraging the model to resolve contradictions and highlight salient differences between $\boldsymbol{x}$ and $\boldsymbol{x}'$ in a principled manner.
We introduce the refinement step because integrating it into the generation pipeline has been shown to enhance the performance and reliability of language models, particularly by fostering self-correction through iterative reasoning~\citep{vernikos2023small,ranaldi2024self,zhang2024small}.

\begin{algorithm}
\caption{MNR pipeline pseudocode}\label{alg:cap}
\begin{algorithmic}
\Require $\boldsymbol{x}$: a factual example, $\boldsymbol{x}'$: a counterfactual example, $p_{draft}$: the draft prompt, $p_{refiner}$: the refiner prompt
\Ensure $\varepsilon$: a narrative explanation for $\left(x, x'\right)$
\State $p_{draft} \gets format\_prompt(p_{draft}, x, x')$
\State $i \gets 0$
\State $E\gets\varnothing$
\While{$i < 3$}
    \State $\varepsilon^i_{draft} \gets \mathcal{M}(p_{draft})$
    \State $E \gets E \cup \varepsilon^i_{draft}$
    \State $i \gets i +1$
\EndWhile
\State $p_{refiner} \gets format\_prompt(p_{refiner}, x, x', E)$
\State $\varepsilon \gets \mathcal{R}(p_{refiner})$
\end{algorithmic}
\end{algorithm}

\paragraph{\textbf{Finetuning Using Knowledge Distillation}}
In order to be more efficient and accurate, our pipeline uses Knowledge Distillation (KD), a training paradigm where a compact student model (an SLM) is trained to replicate the behavior of a powerful teacher model ($T$). This allows us to transfer the powerful reasoning and generation abilities of a large, state-of-the-art model to our much smaller, more efficient SLMs. Since the generation process is split into two different stages, each one addressing a different task, we use KD differently for each one of the stages. 

Concerning the draft generation (stage 1), the KD process begins with creating a high-quality training dataset (See Figure~\ref{fig:dataset-creation}). Given 
a counterfactual explainer $g(\cdot)$, consider a set of $n$ instances $\{\boldsymbol{x}_i\}_{i=1}^n$. For each of these instances, we generate its corresponding optimal counterfactual, i.e., $\boldsymbol{x}_i'= g(\boldsymbol{x}_i)$, thereby obtaining $n$ factual-counterfactual pairs $\{(\boldsymbol{x}_i, \boldsymbol{x}_i')\}_{i=1}^n$. These pairs are embedded into $n$ different structured prompts $p_1,p_2 \dots,p_n$ and then fed to the teacher model $T$ (for further details on the prompts see Figure~\ref{fig:prompts}). The high-quality outputs from $T$ are then used to build a dataset $D_{\mathcal{M}}$ to finetune the DNG.



The KD process is also applied to the refiner $\mathcal{R}$. Here, we leverage the teacher model $T$ to construct a training dataset $D_{\mathcal{R}}$ including not only the factual and counterfactual inputs and prompt, but also a corresponding set of $N=3$ draft explanations. This enriched supervision allows the student model to learn from the implicit variation, redundancy, and conflict present across multiple generated rationales.
Further details on the practical implementation of the KD process can be found in Section~\ref{sec:kd}.

\subsection{Quantitative Evaluation Metrics for Factual-Counterfactual Narratives}
\label{sec:evaluation_metrics}
\begin{figure*}[ht]
    \centering
    \includegraphics[width=1\linewidth]{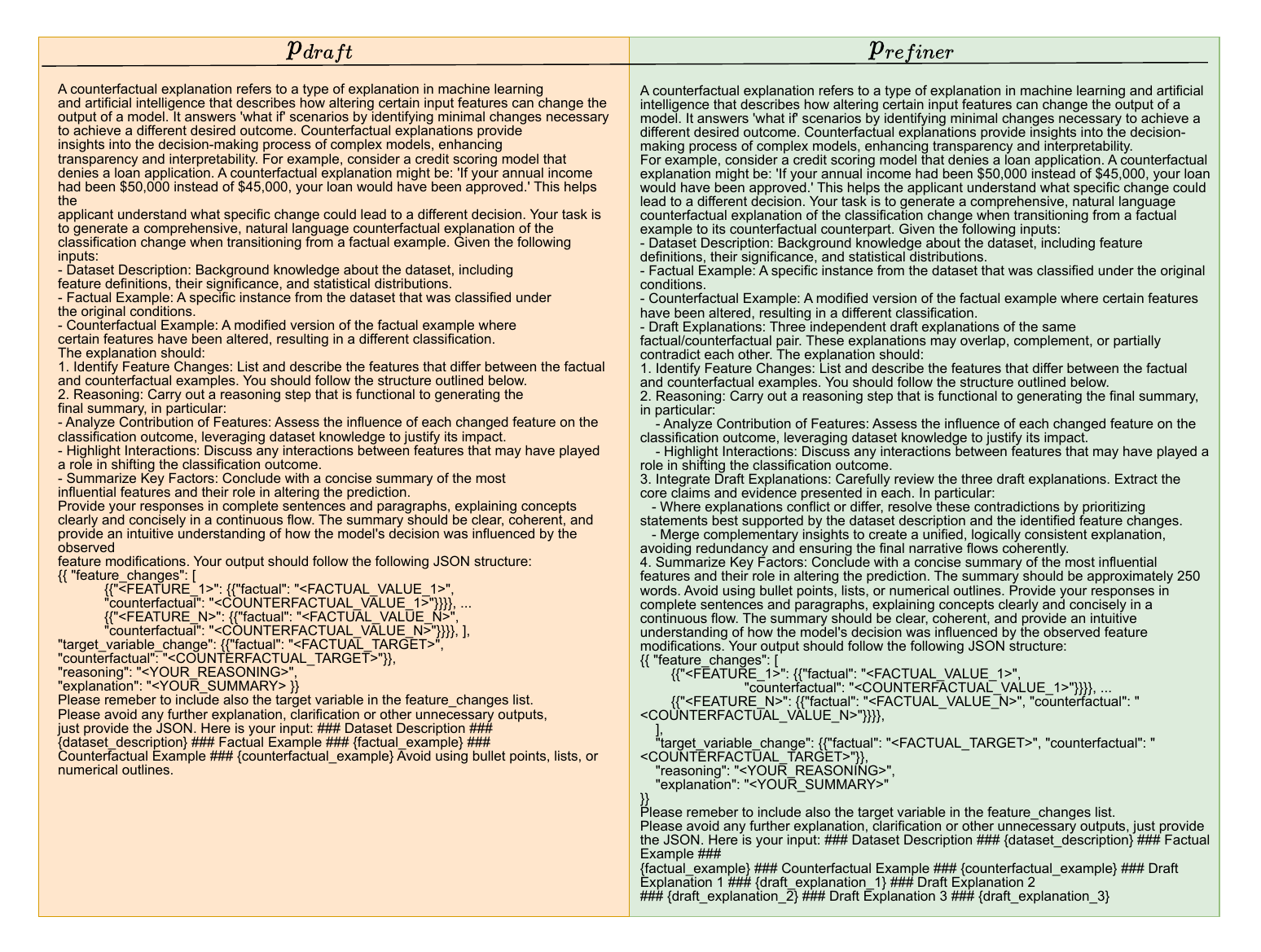}
    \caption{\textbf{Right side:} prompt used to refine multile narratives. \textbf{Left side:} prompt to generate the draft explanations.}
    \label{fig:prompts}
\end{figure*}

To quantitatively assess the narratives produced by our MNR pipeline, we developed a structured evaluation framework that allows us to measure the accuracy of each narrative.
Our evaluation approach, inspired by the methodology in \citep{giorgi2025natural}, requires the model to structure its output. Specifically, the prompt $p$ instructs the SLM to embed its explanation within a structured dictionary format (see Figure~\ref{fig:prompts}). After the refiner model generates the final narrative, we parse this dictionary to extract key information about the feature changes between the factual instance $\boldsymbol{x}$ and its counterfactual $\boldsymbol{x}'$. The core assumption is that a model's ability to populate this structured format correctly reflects its underlying understanding of the narrative. Therefore, errors in the dictionary signal potential inaccuracies in the subsequent natural language narrative. This technique offers a reliable and automated method for evaluating the factual correctness of the generated text. The following metrics quantify how accurately the refined narratives align with the ground truth factual and counterfactual feature changes. 

Given four feature vectors $\mathbf{F}_{fact},~\mathbf{F}_{cf},~\widetilde{\mathbf{F}}_{fact},~ \widetilde{\mathbf{F}}_{cf}$ containing respectively, the factual ground truth, the counterfactual ground truth, the factual values, and the counterfactual values generated by the MNR pipeline, we can define the following metrics:

\noindent\paragraph{\textbf{Average Feature Faithfulness (AvgFF)}} Given the set of all the narratives generated $E$, and all the vectors above for each narrative $\varepsilon$, we can compute the fraction of correctly interpreted features as:
\begin{equation*}
    \text{AvgFF} = \frac{\sum\limits_{\varepsilon\in E}\, \sum\limits_{i=0}^{k} \mathbb{I} \left[ \left( \mathbf{F}^\varepsilon[i]_{fact} = \widetilde{\mathbf{F}}^\varepsilon[i]_{fact} \right) \wedge \left( \mathbf{F}^\varepsilon[i]_{cf} = \widetilde{\mathbf{F}}^\varepsilon[i]_{cf}\right) \right]}{\left| E \right| \cdot k}
\end{equation*}
where $k = |\mathbf{F}_{fact}|$ is the vector size, and $\mathbb{I}$ is the indicator function. 
\noindent\paragraph{\textbf{Perfect Feature Faithfulness (PFF)}} This metric measures the fraction of generated explanations that correctly identify all feature changes. Formally, given the set of all the narratives generated $E$, the \textit{Perfect Feature Faithfulness} is calculated as:

\begin{equation*}
    \text{PFF} = 
    \frac{\sum\limits_{\varepsilon\in E} \mathbb{I}\left[ \left( \mathbf{F}^\varepsilon_{fact} = \widetilde{\mathbf{F}}^\varepsilon_{fact} \right) \wedge \left( \mathbf{F}_{cf}^\varepsilon = \widetilde{\mathbf{F}}^\varepsilon_{cf}\right)\right]}{\left| E \right|}
\end{equation*}

This binary indicator is averaged over all narratives $\varepsilon \in E$, resulting in the proportion of narratives that precisely match all factual and counterfactual feature values.

\paragraph{\textbf{Target Faithfulness (TF)}}
This metric measures the ability of the function $H$ to "understand" the factual and counterfactual dependent variables and compare them against the provided ground truth. Let $E$ be the set of all the narratives generated by our pipeline, $y_{fact}, y_{cf}$ the ground truth target outcomes, and let $\widetilde{y}_{fact}, \widetilde{y}_{cf}$ be the corresponding predicted outcomes by the pipeline. The \emph{Target Faithfulness} computed over all the narratives $\varepsilon \in E$ is defined as:
\begin{equation*}
    \text{TF} =
    \frac{\sum\limits_{\varepsilon\in E} \mathbb{I}\left[ \left( {y}^\varepsilon_{fact} = \widetilde{y}^\varepsilon_{fact} \right) \wedge \left( {y}_{cf}^\varepsilon = \widetilde{{y}}^\varepsilon_{cf}\right)\right]}{\left| E \right|}
\end{equation*}
These metrics provide a clear, quantitative assessment of how reliably the explanations reflect the ground truth provided by the factual and counterfactual scenarios.
\subsection{Narrative Quality}
To evaluate the quality of the generated counterfactual narratives, we employed a questionnaire-based protocol that has been previously used in \cite{giorgi2025natural}. The Narrative Quality dimension focuses on assessing explanations through human evaluation, emphasizing the readability, clarity, and coherence of the provided narrative. The questionnaire comprises 5 specific questions, each rated on a scale from 1 (poor) to 5 (excellent):

\begin{itemize}
    \item Q1: Is the terminology and language used in the narrative appropriate and easy to understand?
    \item Q2: How clear and understandable is the provided narrative regarding feature changes?
    \item Q3: How clearly does the narrative describe the specific feature changes that led to the counterfactual outcome?
    \item Q4: Are the described feature changes easy to interpret, and do they make sense within the context of the original data?
    \item Q5: What is your overall assessment of the clarity and coherence of the narrative?
\end{itemize}

\section{Experiments}
Our experimental framework is detailed below, covering the hardware, datasets, models, and language models used in our evaluation. Experiments are conducted on a system equipped with an AMD Ryzen 9 7900 12-Core Processor, an NVIDIA GeForce RTX 4090 GPU, and 64\,GB of RAM. We evaluate our approach on two benchmark datasets: the Adult dataset~\citep{adult_2}, for predicting whether an individual's income exceeds \$50K/year, and the Titanic dataset~\citep{titanic}, for predicting passenger survival.  The oracle model is a Decision Tree classifier from the \texttt{scikit-learn} library, configured with a \texttt{max\_depth} of 4, a \texttt{criterion} of \texttt{'gini'}, and \texttt{min\_samples\_split} set to 2. This model achieves an accuracy of 0.84 on the Adult dataset and 0.82 on the Titanic dataset. Counterfactual explanations are generated using the DiCE library~\citep{mothilal2020dice}. We utilize models from two distinct families for narrative generation: Qwen2.5~\citep{qwen2.5} and DeepSeek-r1~\citep{deepseekai2025deepseekr1incentivizingreasoningcapability}. The specific hyperparameters used for response generation are detailed in Table~\ref{tab:llm_hyperparams}.

\begin{table}[htbp]
\centering
\caption{Hyperparameter Settings for all the Large Language Models used, namely, Temperature, Top-k, Top-p, Max Tokens, Repetition Penalty}
\setlength{\tabcolsep}{3pt}
\begin{tabular}{lccccc}
\toprule
\textbf{Model Name} & \textbf{T.} & \textbf{Top-k} & \textbf{Top-p} & \textbf{M. T.} & \textbf{R. P.} \\
\midrule
\small Qwen2.5-0.5B-Instruct & 0.6 & 10 & 0.8 & 8192 &1.05  \\
\small Qwen2.5-3B-Instruct & 0.6 & 10 & 0.8 & 8192 & 1.05   \\
\small DeepSeek-R1-D.-Qwen-1.5B & 0.6 & 10 & 0.7 & 8192 & 1.05  \\
\small DeepSeek-R1-D.-Qwen-7B & 0.6 & 10 & 0.7 & 8192 & 1.05   \\
\small DeepSeek-R1-D.-Qwen-32B & 0.7 & 10 & 0.7 & 8192 & 1.05   \\
\bottomrule
\end{tabular}
\label{tab:llm_hyperparams}
\vspace{-5mm}
\end{table}

\subsection{Knowledge Distillation and Fine-tuning}\label{sec:kd}
Since our pipeline uses Knowledge Distillation to improve model performance, we built four new datasets (two for each dataset we tested) tailored explicitly for our tasks (draft narrative generation and narrative refinement). The draft narrative generation dataset consists of a JSON file containing prompt–response pairs, where each prompt is designed for the explanation task and the corresponding response is generated by the \texttt{DeepSeek-R1-Distill-Qwen-32B} teacher model. 
We choose the \texttt{DeepSeek-R1-Distill-Qwen-32B} model due to its strong initial performance in generating coherent and contextually accurate text, especially suited for complex reasoning tasks. For the narrative refinement dataset, we employ fine-tuned \texttt{Qwen2.5-0.5B-Instruct} models as the Draft Narrative Generators, which are the smallest model considered in our evaluation. Our rationale is to construct a dataset using what can be seen as the hypothetically weakest drafter, thereby making the refinement task particularly challenging. In this way, the Refiner is forced to handle noisy, incomplete, or inconsistent drafts, which provides a more rigorous test of its ability to synthesize coherent explanations through self-correction and additional reasoning.
Also in this case, the refiner model that merges the drafts is \sloppy \texttt{DeepSeek-R1-Distill-Qwen-32B}.

Leveraging this structured and targeted dataset, we expect our fine-tuned models to achieve significantly better accuracy and clarity in producing comprehensive and natural-language counterfactual explanations. 
Figure~\ref{fig:dataset-creation} illustrates the dataset generation process for both the draft narrative generation and the narrative refinement tasks. To perform the fine-tuning, we used the Unsloth~\citep{unsloth} framework; all the parameters can be seen in Table~\ref{tab:finetune-hparams}.

\begin{table}[ht] 
\centering 
\caption{Finetuning parameters: Checkpoint: checkpoint used during evaluation, B.S.: Batch Size, T.E.: Training Epochs, M.S.: Max Steps, C.S.S.: Checkpoint Saving Steps} 
    \setlength{\tabcolsep}{3pt}

\begin{tabular}{lccccc} \hline 
Model & Checkpoint & B.S. & T.E. & M.S.  & C.S.S. \\ \hline 
\small Qwen2.5-0.5B-Instruct & 500 & 1 & 1 & 1562  & 50 \\ 

\small Qwen2.5-3B-Instruct & 500 & 1 & 1 & 1562 & 50 \\ 

\small Deepseek-R1-D.-Qwen-1.5B & 400 & 1 & 1 & 1562  & 50 \\ 

\small Deepseek-R1-D.-Qwen-7B & 1000 & 1 & 1 & 1562  & 50 \\

\hline \end{tabular} \label{tab:finetune-hparams} 
\vspace{-1mm}
\end{table}

\section{Results}
In this section, we present the results of our study along two different dimensions. First, we quantitatively and qualitatively analyze the performance of MNR pipeline, assessing both accuracy and narrative quality. Next, we examine the feasibility of employing SLMs for narrative explanation tasks, focusing on efficiency and resource requirements. Finally, we provide an illustrative example to highlight the effectiveness and interpretability of the generated narratives.

\subsection{Multi-Narrative Refinement Results.} 
The results presented in Table~\ref{tab:results-comparison-refiner} provide compelling evidence for the efficacy of our pipeline, highlighting two primary conclusions: the necessity of fine-tuning and the significant performance boost provided by the refinement stage.

Across both datasets, the base models are incapable of performing the narrative generation task, with metrics for Feature Fidelity (FF) and Target Faithfulness (TF) consistently at or near zero. The impact of fine-tuning is dramatic and immediate. For instance, in the ``No Refiner'' baseline for the Adult dataset, the \sloppy \texttt{DeepSeek-R1-Distill-Qwen-1.5B} Draft Explainer sees its Perfect FF score jump from $0.0$ to $0.946$ after fine-tuning. This pattern confirms that distillation is essential to impart the required capabilities to the models.

The value of the two-stage refinement process is most evident when examining the interplay between models. A fine-tuned Draft Explainer alone sets a strong baseline, but adding a capable refiner elevates performance. This is particularly striking when a weaker drafter is used. For example, the \texttt{Qwen2.5-0.5B-Instruct} fine-tuned drafter only achieves a Perfect FF of $0.485$, but when its outputs are processed by the \texttt{Qwen2.5-3B-Instruct} refiner, the score more than doubles $(0.980)$.

The \texttt{Qwen2.5-3B-Instruct} model emerges as the best refiner. Across nearly all configurations on the Adult dataset, it delivers the highest scores, pushing both Perfect FF and TF metrics above $0.97$ and achieving near-zero standard deviation, indicating highly consistent and accurate outputs.

On the Titanic dataset, the best configurations reach a Perfect FF of $0.790$, and the standard deviation is relatively high (often >$0.40$). This suggests that while our pipeline is robust, its performance limit could be influenced by the inherent difficulty of the dataset or by the model's hyperparameters. 

In Table~\ref{tab:results-comparison-refiner}, we also introduced the performance obtained using the teacher model to directly generate the narratives. Results show that our MNR pipeline can perform better than the teacher model used to generate the dataset for the model fine-tuning.

Since the combination draft generator $\mathcal{M}$ Qwen2.5-0.5B-Instruct and refiner $\mathcal{R}$ Qwen2.5-3B-Instruct have the best performance, from now on, we will use these models in our pipeline.
\begin{table}[ht]
    \centering
    \caption{Narrative Quality Evaluation Scores. Our solution (MNR pipeline) has as a draft generator $\mathcal{M}$ Qwen2.5-0.5B-I. and as refiner $\mathcal{R}$ Qwen2.5-3B-I. The model DeepSeek-R1-D.-Qwen-32B is the teacher model we used to build the fine-tuning dataset, Qwen2.5-0.5B-I., instead, is the fine-tuned model without refiner.}
    \setlength{\tabcolsep}{3pt}
    \begin{tabular}{llllll}
        \toprule
        \small \textbf{Model} &  \small \textbf{Q1} &  \small \textbf{Q2} &  \small \textbf{Q3} &  \small \textbf{Q4} &  \small \textbf{Q5} \\
        \midrule
        \footnotesize DeepSeek-R1-D.-Qwen-32B  & \footnotesize 4.5 $\pm$ 0.6 &	\footnotesize 4.4 $\pm$ 0.6 &	\footnotesize 4.3 $\pm$ 0.6 &	\footnotesize 4.2 $\pm$ 1.0 &	\footnotesize 4.4 $\pm$ 0.6\\
        \footnotesize Qwen2.5-0.5B-I. 
        & \footnotesize 3.2 $\pm$ 0.9 &	\footnotesize 3.1 $\pm$ 1.2 &	\footnotesize 3.1 $\pm$ 1.1 &	\footnotesize 2.9 $\pm$ 1.0 &	\footnotesize 3.0 $\pm$	0.7    \\
        \footnotesize MNR pipeline  & \footnotesize 4.4 $\pm$ 0.7 &	\footnotesize 4.6 $\pm$ 0.5 & \footnotesize	4.0 $\pm$ 0.9	& \footnotesize 4.3 $\pm$ 0.6 & \footnotesize	4.4 $\pm$ 0.5\\
        \bottomrule
    \end{tabular}
\label{tab:narrative_quality}
\vspace{-5mm}
\end{table}

\begin{table*}[htbp!]
    \centering
    \caption{Performance comparison of base versus fine-tuned models using the Multi-Narrative Refinement (MNR) pipeline. The first row for each dataset shows the performance of the teacher model solving the counterfactual narrative generation problem by itself. The rows with \emph{No Refiner} serve as a baseline to get the model performance without a refiner attached. The results shown here evaluate the impact of fine-tuning the Draft Explainer model directly. If the Refiner is indicated, the rows evaluate the performance of the Refiner model. In these cases, the comparison is between a plain and a fine-tuned Refiner, both of which use drafts generated by an already fine-tuned Draft Explainer.}
    \label{tab:results-comparison-refiner}

    \newcommand{\teacher}{DeepSeek-Q-32B}
    \newcommand{\qwenA}{Qwen2.5-0.5B-I}
    \newcommand{\dsqA}{DeepSeek-Q-1.5B}
    \newcommand{\qwenB}{Qwen2.5-3B-I}
    \newcommand{\dsqB}{DeepSeek-Q-7B}

    \begin{tabular}{
        l 
        l 
        l 
        l 
        l   
        l   
        l   
        l   
    }
    \toprule
    \multirow{2}{*}{\textbf{Draft Model ($\mathcal{M}$)}} & \multirow{2}{*}{\textbf{Refiner Model ($\mathcal{R}$)}} & \multicolumn{2}{c}{\textbf{AvgFF $\pm$ std} $\uparrow$} & \multicolumn{2}{c}{\textbf{PFF} $\uparrow$} & \multicolumn{2}{c}{\textbf{TF} $\uparrow$} \\
    \cmidrule(lr){3-4} \cmidrule(lr){5-6} \cmidrule(lr){7-8}
    & & {\textbf{Base}} & {\textbf{Tuned}} & {\textbf{Base}} & {\textbf{Tuned}} & {\textbf{Base}} & {\textbf{Tuned}} \\
    \midrule
    \multicolumn{8}{l}{\textbf{Adult Dataset}} \\
    \midrule
    \multicolumn{2}{c}{\teacher~(\textit{Teacher})} & \multicolumn{2}{c}{0.945$\pm$0.10} & \multicolumn{2}{c}{0.865} & \multicolumn{2}{c}{0.980} \\
    \cmidrule(lr){1-8}
    \multirow{5}{*}{\qwenA} & \textit{No Refiner} & 0.000 $\pm$ {\text{n.d.}} & 0.485 $\pm$ 0.24 & 0.000 & 0.485 & 0.000 & 0.515 \\
    & \qwenA & 0.012 $\pm$ 0.19 & 0.796 $\pm$ 0.12 & 0.005 & 0.790 & 0.000 & 0.810 \\
    & \dsqA & 0.000 $\pm$ {\text{n.d.}} & \underline{0.868 $\pm$ 0.10} & 0.000 & \underline{0.855} & 0.150 & 0.750 \\
    & \qwenB & 0.122 $\pm$ 0.10 & \textbf{0.980 $\pm$ 0.00} & 0.120 & \textbf{0.980} & 0.125 & \textbf{0.980} \\
    & \dsqB & 0.048 $\pm$ 0.44 & 0.854 $\pm$ 0.24 & 0.035 & 0.755 & 0.120 & \underline{0.965} \\
    \cmidrule(lr){1-8}
    \multirow{5}{*}{\dsqA} & \textit{No Refiner} & 0.000 $\pm$ {\text{n.d.}} & \underline{0.946 $\pm$ 0.03} & 0.000 & \underline{0.945} & 0.135 & 0.410 \\
    & \qwenA & 0.005 $\pm$ {\text{n.d.}} & 0.892 $\pm$ 0.08 & 0.005 & 0.890 & 0.000 & 0.900 \\
    & \dsqA & 0.000 $\pm$ {\text{n.d.}} & 0.902 $\pm$ 0.10 & 0.000 & 0.885 & 0.120 & 0.785 \\
    & \qwenB & 0.085 $\pm$ 0.00 & \textbf{0.976 $\pm$ 0.04} & 0.085 & \textbf{0.970} & 0.085 & \textbf{0.980} \\
    & \dsqB & 0.030 $\pm$ 0.51 & 0.823 $\pm$ 0.26 & 0.030 & 0.700 & 0.070 & \underline{0.955} \\
    \cmidrule(lr){1-8}
    \multirow{5}{*}{\qwenB} & \textit{No Refiner} & 0.053 $\pm$ 0.10 & \underline{0.988 $\pm$ 0.00} & 0.050 & \underline{0.988} & 0.055 & \underline{0.988} \\
    & \qwenA & 0.007 $\pm$ 0.42 & 0.908 $\pm$ 0.02 & 0.005 & 0.905 & 0.005 & 0.910 \\
    & \dsqA & 0.005 $\pm$ 0.13 & 0.922 $\pm$ 0.08 & 0.005 & 0.905 & 0.090 & 0.795 \\
    & \qwenB & 0.085 $\pm$ 0.00 & \textbf{0.995 $\pm$ 0.00} & 0.085 & \textbf{0.995} & 0.085 & \textbf{0.995} \\
    & \dsqB & 0.038 $\pm$ 0.48 & 0.840 $\pm$ 0.24 & 0.035 & 0.750 & 0.105 & 0.945 \\
    \cmidrule(lr){1-8}
    \multirow{5}{*}{\dsqB} & \textit{No Refiner} & 0.251 $\pm$ 0.47 & 0.855 $\pm$ 0.11 & 0.230 & 0.855 & 0.405 & 0.865 \\
    & \qwenA & 0.005 $\pm$ {\text{n.d.}} & 0.873 $\pm$ 0.10 & 0.005 & 0.865 & 0.000 & 0.885 \\
    & \dsqA & 0.000 $\pm$ {\text{n.d.}} & \underline{0.904 $\pm$ 0.05} & 0.000 & \underline{0.895} & 0.106 & 0.770 \\
    & \qwenB & 0.080 $\pm$ 0.00 & \textbf{0.973 $\pm$ 0.04} & 0.080 & \textbf{0.970} & 0.080 & \underline{0.975} \\
    & \dsqB & 0.035 $\pm$ 0.47 & 0.843 $\pm$ 0.26 & 0.030 & 0.715 & 0.090 & \textbf{0.980} \\
    \midrule
    \multicolumn{8}{l}{\textbf{Titanic Dataset}} \\
    \midrule
    \multicolumn{2}{c}{\teacher~(\textit{Teacher})} & \multicolumn{2}{c}{0.608 $\pm$ 0.47} & \multicolumn{2}{c}{0.590} & \multicolumn{2}{c}{0.945} \\
    \cmidrule(lr){1-8}
    \multirow{5}{*}{\qwenA} & \textit{No Refiner} & 0.000 $\pm$ {\text{n.d.}} & \textbf{0.790 $\pm$ 0.16} & 0.000 & \underline{0.740} & 0.000 & 0.840 \\
    & \qwenA & 0.052 $\pm$ 0.40 & 0.251 $\pm$ 0.49 & 0.040 & 0.243 & 0.015 & 0.592 \\
    & \dsqA & 0.000 $\pm$ {\text{n.d.}} & 0.522 $\pm$ 0.47 & 0.000 & 0.510 & 0.050 & 0.805 \\
    & \qwenB & 0.040 $\pm$ 0.00 & \underline{0.782 $\pm$ 0.40} & 0.040 & \textbf{0.780} & 0.040 & \textbf{0.985} \\
    & \dsqB & 0.080 $\pm$ 0.48 & 0.562 $\pm$ 0.49 & 0.075 & 0.550 & 0.210 & \underline{0.960} \\
    \cmidrule(lr){1-8}
    \multirow{5}{*}{\dsqA} & \textit{No Refiner} & 0.003 $\pm$ 0.15 & 0.475 $\pm$ 0.49 & 0.000 & 0.470 & 0.035 & 0.095 \\
    & \qwenA & 0.052 $\pm$ 0.38 & 0.350 $\pm$ 0.49 & 0.045 & 0.340 & 0.000 & 0.660 \\
    & \dsqA & 0.000 $\pm$ {\text{n.d.}} & 0.507 $\pm$ 0.48 & 0.000 & 0.500 & 0.030 & 0.790 \\
    & \qwenB & 0.050 $\pm$ 0.00 & \textbf{0.790 $\pm$ 0.40} & 0.050 & \textbf{0.785} & 0.050 & \textbf{0.995} \\
    & \dsqB & 0.117 $\pm$ 0.49 & \underline{0.568 $\pm$ 0.49} & 0.110 & \underline{0.555} & 0.240 & \underline{0.955} \\
    \cmidrule(lr){1-8}
    \multirow{5}{*}{\qwenB} & \textit{No Refiner} & 0.000 $\pm$ {\text{n.d.}} & \underline{0.635 $\pm$ 0.46} & 0.000 & \underline{0.625} & 0.005 & 0.920 \\
    & \qwenA & 0.040 $\pm$ 0.34 & 0.383 $\pm$ 0.49 & 0.030 & 0.365 & 0.010 & 0.715 \\
    & \dsqA & 0.005 $\pm$ 0.22 & 0.507 $\pm$ 0.48 & 0.005 & 0.495 & 0.055 & 0.840 \\
    & \qwenB & 0.065 $\pm$ 0.00 & \textbf{0.738 $\pm$ 0.44} & 0.065 & \textbf{0.737} & 0.065 & \textbf{0.987} \\
    & \dsqB & 0.102 $\pm$ 0.49 & 0.578 $\pm$ 0.49 & 0.095 & 0.565 & 0.210 & \underline{0.975} \\
    \cmidrule(lr){1-8}
    \multirow{5}{*}{\dsqB} & \textit{No Refiner} & 0.287 $\pm$ 0.36 & 0.385 $\pm$ 0.49 & 0.280 & 0.280 & 0.345 & 0.490 \\
    & \qwenA & 0.033 $\pm$ 0.37 & 0.417 $\pm$ 0.49 & 0.030 & 0.410 & 0.005 & 0.720 \\
    & \dsqA & 0.010 $\pm$ 0.39 & 0.522 $\pm$ 0.47 & 0.010 & 0.515 & 0.035 & \underline{0.805} \\
    & \qwenB & 0.050 $\pm$ 0.00 & \textbf{0.777 $\pm$ 0.41} & 0.050 & \textbf{0.775} & 0.050 & \textbf{0.990} \\
    & \dsqB & 0.133 $\pm$ 0.48 & \underline{0.583 $\pm$ 0.49} & 0.120 & \underline{0.570} & 0.245 & \textbf{0.990} \\
    \bottomrule
    \end{tabular}
\end{table*}

\subsection{Qualitative Results}
The survey has been conducted on a sample of 15 people with different backgrounds, ages, and genders. The results of the human evaluation in Table \ref{tab:narrative_quality} highlight the significant impact of our pipeline on the narrative quality. The teacher model (DeepSeek-R1-D.-Qwen-32B) gets high scores across all five of the evaluation criteria.
Notably, our approach (MNR pipeline) achieves comparably high scores in all categories, with Q4 (interpretability and contextual coherence) receiving the highest rating of 4.3 (slightly higher than the teacher). This suggests that the narratives generated by the MNR pipeline are as clear as the one generated by the 32B model used as teacher. Similarly, Q2 and Q3, which assess the clarity and specificity of feature changes, exhibit strong performance (4.6 and 4.0, respectively), indicating that our solution effectively conveys how and why changes lead to a counterfactual outcome.

In contrast, the model without refiner scores significantly lower, particularly in Q4 and Q5, with ratings of 2.9 and 3.0, respectively. These results suggest that without the refinement step, the model struggles to produce clear and coherent narratives, making them harder to interpret and understand.

\subsection{Feasibility}\label{sec:feasibility}
Beyond accuracy and quality, the feasibility of deploying counterfactual narrative generators critically depends on their computational requirements. In our study, we compare three settings: a teacher LLM, a worker SLM without refinement, and a Multi-Narrative Refinement pipeline composed of two SLMs, both finetuned. To assess their practical applicability, we analyze three key dimensions: (i) model size, (ii) inference time, and (iii) energy consumption during explanation generation.

\begin{figure}
    \centering
    \includegraphics[width=1\linewidth]{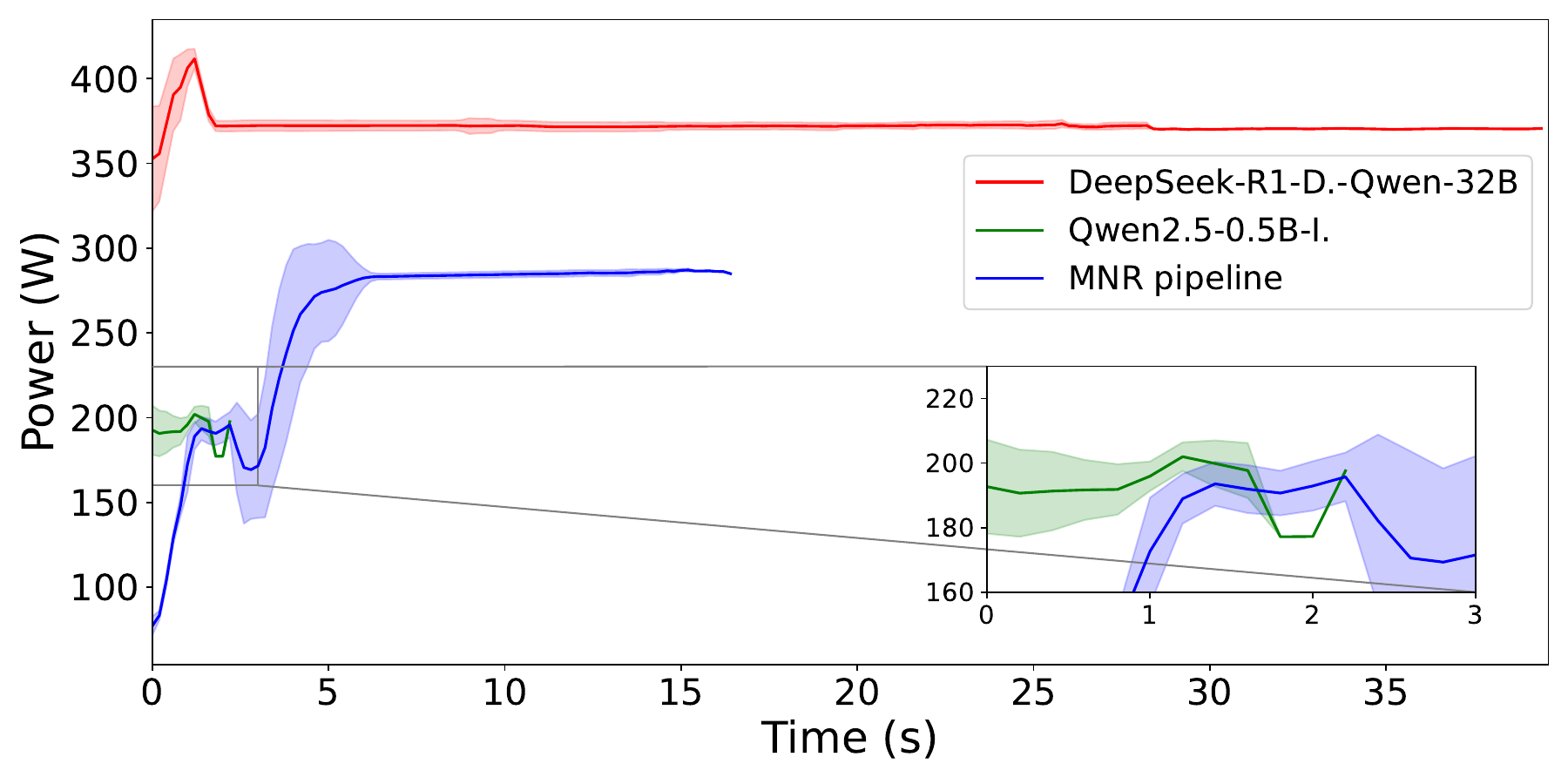}
    \caption{GPU power comparison of three techniques for narrative generation. 
    Our approach (MNR pipeline) uses Qwen2.5-0.5B-I. as draft generator $\mathcal{M}$ and Qwen2.5-3B-I. as refiner $\mathcal{R}$. 
    DeepSeek-R1-D.-Qwen-32B is the teacher model, while Qwen2.5-0.5B-I is the fine-tuned model without refiner. 
    Solid lines show average power; shaded areas denote standard deviation.}
    \label{fig:power}
\end{figure}

Model size directly affects deployability. Large LLMs demand tens of gigabytes of memory, whereas distilled SLMs are an order of magnitude smaller, enabling deployment on commodity hardware. In the case of the MNR pipeline, model size reflects the combined size of both SLMs. Regarding inference time, the MNR pipeline’s reported latency corresponds to the full generation process of a final explanation, which includes producing three independent draft narratives followed by a refinement step. By contrast, the SLM without refinement and the teacher model $T$ each generate only a single draft.
For energy consumption, we measure the total GPU energy required to generate a single explanation. GPU power draw was sampled every 200 milliseconds, yielding a fine-grained trace of instantaneous consumption. The total energy was then computed by integrating power measurements over time. Specifically, given the set of samples
$
S = \{(t_0,P_0), (t_1,P_1), \ldots, (t_n,P_n)\},
$
where $t_i$ denotes the timestamp of sample $i$ and $P_i$ the corresponding power (in Watts), the energy consumption is approximated as
$$
E_{total} \approx \sum_{i=1}^{n} \frac{P_{i-1} + P_i}{2} \cdot (t_i - t_{i-1}).
$$
This calculation yields the total energy expenditure (in Joules) for generating a single explanation under each configuration. 
Table~\ref{tab:feasibility} reports the results for the Adult dataset, comparing the LLM, the worker SLM without refiner, and the MNR pipeline.
Our findings show that the MNR approach reduces memory usage by a factor of 2.7 compared to the large model (\texttt{DeepSeek-R1-D.-Qwen-32B}), while also lowering inference time by nearly 50\% and energy consumption by 62\% (see Fig.~\ref{fig:power} for the power consumption trends over time). Although the standalone SLM achieves even lower memory usage, faster inference, and reduced energy consumption, its performance is consistently halved across all evaluation metrics.

\begin{table}[htpb]
\centering
\caption{Feasibility results for explanation generation across different pipelines for the Adult dataset. 
The first row reports the performance of the teacher T, the second corresponds to the worker SLM without refinement, and the third to the MNR pipeline composed of two fine-tuned SLMs, a draft generator $\mathcal{M}$ Qwen2.5-0.5B-I. and a refiner $\mathcal{R}$ Qwen2.5-3B-I. 
}
\begin{tabular}{lcccc}
\toprule
\textbf{\small Model Name} & \textbf{\small Size (GB)} & \textbf{\small Time (s)}
& \textbf{\small Energy (J)} \\
\midrule
\footnotesize DeepSeek-R1-D.-Qwen-32B &
\small 18.16 &
\small 20.5 $\pm$ 4.1 &
\small 7613.6 $\pm$ 1536.5   \\

\footnotesize Qwen2.5-0.5B-I. &
\small 0.92 &
\small 1.23 $\pm$ 0.29 &
\small 215.5 $\pm$ 53.3    \\

\footnotesize MNR pipeline &
\small 6.71 &
\small 12.33 $\pm$ 2.0 &
\small 2867.2 $\pm$ 548.4   \\
\bottomrule
\end{tabular}

\label{tab:feasibility}
\vspace{-5mm}
\end{table}

\subsection{Narrative Example} Table~\ref{tab:reasoning} illustrates the effectiveness of the MNR pipeline by comparing the raw drafts generated by the Draft Narrative Generator $\mathcal{M}$ with the merging phase produced by the Refiner $\mathcal{R}$.

The three drafts provide diverse but imperfect accounts: for instance, Draft 1 correctly identifies occupation and marital status as influential, yet introduces minor ambiguities by overstating the role of race; Draft 2 emphasizes the importance of the first two feature changes, and also speculates that race change does not have a direct impact on prediction; Draft 3, instead, misattributes relevance to unchanged factors such as age, thereby introducing noise into the explanation. Left on their own, each draft offers only a partial and potentially misleading picture of the prediction shift.

The refinement stage addresses these limitations by systematically comparing the drafts, resolving contradictions, and merging complementary insights. In the merging phase shown in the table, the Refiner explicitly discards Draft 3’s error (“Draft 3 seems incorrect because age didn’t change”) and aligns the narrative with the factual–counterfactual pair. At the same time, it strengthens the explanation by integrating overlapping claims from Drafts 1 and 2: namely, that occupation is the primary driver of the income change, while marital status contributes additional financial stability. The Refiner also goes beyond simple aggregation, providing deeper reasoning on feature interactions — for example, noting that being married and working in a white-collar job together exert a stronger effect than either factor alone, or that the impact of the change in race from `White`to `Other` is uncertain or secondary.

\begin{table*}[ht!]
    \centering
    \caption{Comparison between the three draft explanations and the Refiner’s reasoning phase for the Adult dataset.}
    \begin{tabular}{c|c|c|c}
        \midrule
        \multicolumn{4}{c}{\Large \textbf{Example}} 
        \\
        \hline
        \multicolumn{4}{p{0.95\textwidth}}{
        \footnotesize
            \textbf{Factual}: age: 33; workclass: Private; education: School; marital status: Divorced; occupation: Blue-Collar; race: White; gender: Male; hours per week: 35 \par
            \textbf{Counterfactual}: age: 33; workclass: Private; education: School; marital status: \emph{Married}; occupation: \emph{White-Collar}; race: \emph{Other}; gender: Male; hours per week: 35
        }
        \\
        \hline
        \multicolumn{2}{c}{
        DNG ($\mathcal{M}$): \textbf{Qwen 2.5 0.5B}} & \multicolumn{2}{c}{Refiner ($\mathcal{R}$): \textbf{Qwen 2.5 3B}}
        \\
        \hline
        \multicolumn{2}{p{5cm}}{\footnotesize  \textbf{Draft 1}: The most influential features contributing to the change in income prediction are marital status, occupation, and race. Being married, which often signifies stability, likely enhances earning potential. Changing occupation to white-collar suggests a higher-paying job, which directly impacts income. Lastly, race may play a role in the model's perception of economic status, possibly reflecting broader societal factors. Together, these changes shift the model's assessment. \textbf{Draft 2}: The most influential factor in altering the prediction is the occupation change from 'Blue-Collar' to 'White-Collar'. This shift suggests a move to a higher-paying, professional role, which is strongly correlated with higher income. Additionally, the marital status change to 'Married' may imply financial stability or dual income, contributing to the higher income prediction. The race change might reflect broader economic disparities, though this effect is less direct. \textbf{Draft 3}: The increase in income classification from 'Income <=50K' to 'Income >50K' is primarily driven by three factors: older age, married status, and a white-collar job. These changes suggest that individuals in these categories are more likely to earn above 50K} & \multicolumn{2}{p{11cm}}{\footnotesize \textbf{Merging Phase:} Okay, so I need to help the user by generating a comprehensive counterfactual explanation based on the given inputs. Let me start by understanding the problem. First, the dataset is the Adult dataset, which predicts income above or below 50K. The factual example has an income of 0 (<=50K), and the counterfactual has 1 (>50K). So, the goal is to explain why changing certain features caused this shift. Looking at the factual and counterfactual examples: Factual: - age: 33 - workclass: Private - education: School - marital\_status: Divorced - occupation: Blue-Collar - race: White - gender: Male - hours-per-week: 35 Counterfactual: - age: 33 - workclass: Private - education: School - marital\_status: Married - occupation: White-Collar - race: Other - gender: Male - hours-per-week: 35. So, the features that changed are: - marital\_status: Divorced -> Married - occupation: Blue-Collar -> White-Collar - race: White -> Other. Wait, but looking back, the factual example's education is 'School', while the counterfactual's is still 'School'? Or did I misread? Oh no, wait, the factual example's education is 'School', and the counterfactual's is 'School' too. So those didn't change. Hmm, maybe I missed something. Let me check again. No, both have 'School' as education. So the only changes are marital\_status, occupation, and race. Wait, in the factual example, the occupation is 'Blue-Collar', and in the counterfactual, it's 'White-Collar'. Also, race changes from 'White' to 'Other'. Now, looking at the draft explanations: Draft 1 says marital status, occupation, and race are the main factors. It mentions being married enhances earning potential, white-collar jobs indicate higher pay, and race reflects economic status. Draft 2 focuses mainly on occupation change to white-collar, suggesting higher income correlation. It also mentions marital status implying financial stability and possible dual income. Race might reflect economic disparities. Draft 3 talks about older age, married status, and white-collar job driving the income increase. But in our case, age remains 33, so that doesn't apply here. Hmm, so Draft 3 seems incorrect because age didn't change. So we can disregard that part. Now, analyzing the contributions: Occupation is a big one. Blue-Collar typically pays less than White-Collar. So changing to White-Collar likely increases income. Marital status: Married might correlate with higher income due to potential dual income or stability. Race: Changing from White to Other this is tricky. The dataset's race categories might have different income implications. Maybe 'Other' includes groups with higher average incomes, but it's not straightforward. But in the dataset description, race is a categorical variable, and the target variable is income. So perhaps 'Other' isn't necessarily worse; maybe it's neutral or positive in this context. Interactions: Maybe being married and having a White-Collar job together have a stronger effect than individually. Integrating the drafts: Draft 1 and 2 agree on occupation and marital status. Draft 2 adds marital status implying financial stability. Draft 3 incorrectly mentions age and doesn't address all changes. So, the key factors are occupation and marital status. Race's impact is uncertain but might contribute positively. Putting it all together, the explanation should highlight occupation as the primary driver, followed by marital status, and note the race change without overemphasizing it since age remained the same. Narrative: Occupation change from Blue-Collar to White-Collar likely increased income. Marital status from Divorced to Married might imply stability or dual income. Race change to Other might have a slight positive effect. The main factors are occupation and marital status. The race change is less impactful but still contributes. Wait, but in the factual example, the person is male, divorced, working blue-collar, white, and making <=50K. In the counterfactual, they're married, white-collar, male, and making >50K. So the key changes are occupation and marital status. So, the summary should focus on those two features, noting that occupation is a strong indicator of income level, and marital status can add to that effect. Race's role is secondary or unclear without more data. I think I've got it.} \\
        \hline
    \end{tabular}
    \label{tab:reasoning}
\end{table*}

This process exemplifies the self-correcting nature of MNR: by leveraging multiple imperfect narratives, the pipeline constructs a coherent, accurate, and more nuanced explanation than any single draft could provide. As a result, the final narrative not only corrects factual inconsistencies but also enriches the reasoning, highlighting feature interactions and causal plausibility in a way that enhances both fidelity and user interpretability.
For other examples, please refer to the code \url{https://github.com/flaat/llm_kd}.

\section{Conclusion}
In this work, we introduced the Multi-Narrative Refinement pipeline, a novel approach to generating human-readable counterfactual narratives from tabular data. By combining knowledge distillation with a structured two-stage draft-and-refine process, we demonstrated that Small Language Models can be empowered to deliver coherent and contextually accurate narratives that demystify the decisions for complex AI models.

Our experiments yield several key insights. First, fine-tuning via knowledge distillation is not merely advantageous but indispensable: base SLMs without distillation fail to perform the task, whereas distilled versions consistently achieve high feature fidelity and narrative coherence. Second, the refinement stage consistently amplifies performance by resolving contradictions between drafts, correcting errors, and synthesizing more robust explanations. 

Taken together, these results highlight the dual importance of distillation and multi-narrative refinement. The MNR pipeline not only advances the state of the art in counterfactual narrative generation but also demonstrates a practical path toward deploying smaller, resource-efficient models that preserve both reasoning depth and explanatory quality.

\section{Social Impact}
The research presented in this paper carries relevant social implications by addressing the need for transparency and interpretability in AI systems.
The positive impact of our work is the democratization of AI explainability. By converting technical counterfactuals into clear and accessible natural language narratives, our framework empowers individuals who are not AI experts to understand the reasoning behind automated decisions that directly affect their lives. This fosters greater trust in AI systems and provides a basis for users to contest unfair or erroneous outcomes. Furthermore, in high-stakes domains like finance and healthcare, our method can help organizations meet regulatory requirements for transparency, such as those outlined in the EU AI Act (see Section~\ref{sec:intro}). Furthermore, by enabling the use of smaller and more efficient models, our work also contributes to a more sustainable and economically viable approach to deploying explainable AI (see Section~\ref{sec:feasibility}), making these powerful tools available to a wider range of organizations.
Despite its benefits, this technology is not without risks. For example, the generated narratives could be highly persuasive but subtly inaccurate and could mask underlying model flaws or biases \citep{berti2025emergent, chen2023agentverse, hagendorff2024deception}. 
The process of simplification inherent in creating a narrative might also omit crucial details, creating a gap between the explanation's perceived meaning and the model's true internal logic.
\begin{acks}
This work was partially supported by the following projects: SERICS (PE00000014) under the National Recovery and Resilience Plan funded by the European Union NextGenerationEU; HypeKG – Hybrid Prediction and Explanation with Knowledge Graphs (H53D23003720006) funded by the Italian Ministry of University and Research under the PRIN 2022 program; GHOST – Protecting User Privacy from Community Detection in Social Networks (B83C24007070005) funded by Sapienza University of Rome - "Progetti di Ricerca Grandi."
\end{acks}

\bibliographystyle{ACM-Reference-Format}
\bibliography{sample-base}

\end{document}